\documentclass[11pt]{article}
\usepackage[table]{xcolor}
\definecolor{ours}{RGB}{222,235,247}
\definecolor{anomcolor}{RGB}{222,235,247}   
\definecolor{attkcolor}{RGB}{253,230,205}   
\newcommand{\takeaway}[1]{\par\medskip\noindent
  \fcolorbox{black!40}{ours}{\parbox{\dimexpr\linewidth-2\fboxsep-2\fboxrule\relax}{\textbf{Key takeaway.} #1}}\par\medskip}
\usepackage[margin=1in]{geometry}
\usepackage{amsmath,amssymb}
\usepackage{array,booktabs,tabularx}
\usepackage{float}
\usepackage{graphicx}
\usepackage{microtype}
\usepackage[numbers,sort&compress]{natbib}
\usepackage{xcolor}
\usepackage[hidelinks]{hyperref}
\usepackage{url}

\newcommand{\kev}{Kev}
\newcommand{\jev}{Jev}
\newcommand{\msdecision}{Microsoft Decision-1}
\newcommand{\oaidecision}{OpenAI Decisions}
\newcommand{\method}{Lev-3B}
\newcommand{\blank}{\phantom{00.000}}

\newcommand{\softmax}{\operatorname{softmax}}

\newcommand{\argmin}{\operatorname*{arg\,min}}
\newcommand{\E}{\operatorname{E}}

\newcolumntype{Y}{>{\raggedright\arraybackslash}X}
\title{\method{}: A Data Recipe for Compact Probabilistic Log Decision Models}

\author{
  Leo Betthauser\thanks{Equal contribution.},
   Yingqi Liu\footnotemark[1],
   Uday Shah\footnotemark[1],
   Jishang Wei,
   Sainyam Galhotra,
   Liang Gou,\\
   Vedant Dharnidharka,
   Hao Yang \\
  Cisco\\
 \small{ \url{leobetthauser@gmail.com}, \url{{yingqliu, udshah, jiswei, sgalhotr, lgou, vdharnid, haoya2}@cisco.com}}
}

\begin{document}
\maketitle

\begin{abstract}
Many log-analysis tasks involve choosing among known options, such as detecting anomalous sessions or identifying attack techniques, and benefit from calibrated probabilities. We present \method{}, an open 3B-parameter decision model built on Granite-4.1-3B-Base with a shared pointer readout, and study its training recipe. The recipe has two key parts. First, training follows a curriculum: the model learns bounded natural-language decisions, using no log data, before it sees any logs. Second, the log stage mixes the target tasks with general log data spanning 35 auxiliary tasks unrelated to either evaluation task. We evaluate one shared checkpoint on sources seen in training and on sources excluded from training, calibration and model selection. The two-stage recipe transfers substantially better than training on log data alone. Adding general log data improves performance on excluded sources. At a fraction of their serving cost, \method{} outperforms prompted frontier models on anomaly detection, including on an excluded logging system, but trails them on attack-technique classification. Its probabilities are well calibrated on familiar sources, but on the excluded logging system it confidently misses anomalies. We release the checkpoint, recipe, data manifests, and prediction-level artifacts.
\end{abstract}

\section{Introduction}
System and security logs combine natural-language messages with identifiers,
repeated templates, numerical values, and temporal order. Many downstream tasks
nonetheless reduce to a choice among a small set of known alternatives, such as
normal versus anomalous~\citep{xu2009detecting,lou2010invariants,lin2016logcluster,du2017deeplog,meng2019loganomaly,zhang2019logrobust, le2021neurallog,guo2021logbert,guan2024logllm}, or one technique from a fixed catalog~\citep{milajerdi2019holmes,hassan2020rapsheet,strom2018attack,niloy2026multisource,husari2017ttpdrill}. Such decisions
can be cast as prediction over a finite, explicitly described option set, with
the model returning a probability distribution over the options rather than
generated text.

Two properties make these decisions hard to learn and to transfer. First,
logging systems differ in message templates, vocabularies, and field
conventions, so a detector fitted to one system may degrade on another~\citep{chen2020logtransfer,han2021logtad,le2022howfar,landauer2024critical}. Second, anomalies are rare. A small
false-positive rate can dominate the alert stream~\citep{axelsson2000base}, and precision-based metrics
vary with evaluation prevalence. A model can therefore score well on the
systems it was trained on and still make confident errors on an unfamiliar one.
Accuracy on familiar sources alone does not show how a model behaves when the
source changes.

Prompting a frontier model avoids training but fits this setting poorly. Logs
are high-volume and often sensitive, which favors a model that can be
self-hosted cheaply. Prompted models also return labels rather than
probability vectors, so their outputs cannot be thresholded, ranked, or
calibrated for alert triage. Typed decision models address the output side:
\jev{} maps unstructured state and typed questions directly to probabilistic
decisions \cite{typesafe2026jev}, and the open \kev{} implementation provides a trainable
approximation of this interface \cite{palmer2026kev}. What remains open for logs is the
\emph{data recipe}: which data to train on, in what order, and how to keep
held-out sources out of every adaptation step. We therefore ask:

\begin{quote}
\emph{Can a compact, open decision model be adapted to make accurate,
well-calibrated decisions on heterogeneous logs, including logging systems
excluded from adaptation, and which data in the recipe account for its
performance?}
\end{quote}

We answer this question with Lev-3B, built on Granite-4.1-3B-Base with a shared
pointer readout, and trained in two stages. The first stage teaches the model
bounded natural-language decisions under a common option-based interface,
using no log data. The second stage continues from that checkpoint on a mixture
of three log data sources: anomaly detection on HPC logs, ATT\&CK
parent-technique classification with synthetic augmentation of scarce security
labels, and \emph{general log data}.
General log data is not tied to either evaluation task; its role is to expose
the model to the variety of log formats and to help it carry what it learns to
logging systems it has not seen. A single shared checkpoint serves both tasks,
so the evaluation measures both what specialization adds and what is lost when
the tasks are combined.

We evaluate the checkpoint on represented sources (BGL, Spirit, Thunderbird,
and seen ATT\&CK sourcetypes) and on sources excluded from every training,
calibration, and model-selection step (HDFS and 18 withheld ATT\&CK
sourcetypes). Because decision models expose probabilities, we evaluate those
probabilities as well as accuracy: raw outputs, a single post-hoc temperature
fitted on independent non-target data, and confident errors. We organize the
study around three questions: how Lev-3B compares with hosted and frontier
models, and which recipe components matter (RQ1); whether general log data
improves transfer to excluded sources (RQ2); and whether the predicted
probabilities are reliable (RQ3). We summarize our key findings below.
\begin{itemize}
  \item \textbf{Adaptation, not prompting, drives log performance.} Lev-3B
  reaches 88--99 anomaly F1 on the represented HPC sources, where prompted
  frontier models stay at or below 28 on Spirit and 10 on Thunderbird even
  with five demonstrations. On the excluded HDFS source it has the best F1 of
  all systems (43.3 vs.\ at most 25.9) and roughly four times \jev{}'s AUPRC.
  On ATT\&CK it outperforms the three hosted decision models (\jev{}, \msdecision{}, and \oaidecision{}) but trails prompted GPT-5.6, especially on withheld sourcetypes.
  \item \textbf{General log data drives transfer.} Adding ATT\&CK supervision
  costs anomaly performance (Spirit F1 91.8 $\rightarrow$ 77.5). Adding general
  log data recovers most of that loss (88.3) and improves both excluded sources:
  HDFS F1 rises from 25.3 to 43.3, AUPRC from 0.25 to 0.41, and withheld
  ATT\&CK accuracy from 34.3 to 42.1.
\item \textbf{Probabilities are mostly reliable on familiar sources, but miscalibrated when the data source changes.} 
  Raw Lev-3B probabilities are close to calibrated on represented
  anomaly sources and have lower reported NLL than those of all three hosted decision models on both excluded sources. On HDFS,
  however, nearly all of its confident errors are missed anomalies (2{,}105
  vs.\ 84 false alarms). A fitted temperature removes almost all confident false alarms but leaves most confident misses.
\end{itemize}

This report makes the following contributions.

\begin{enumerate}
  \item \textbf{A data recipe for log decisions.} We combine natural-language
  decision training with a joint mixture of anomaly data, ATT\&CK data with
  synthetic augmentation, and general log data, behind one shared pointer
  readout, yielding one compact open checkpoint for both tasks.
  \item \textbf{Controlled ablations showing what each data source
  contributes.} Ablations that add one data subset at a time show that task
  data determines task skill, that combining tasks causes interference, and
  that general log data offsets most of that interference while improving
  transfer to excluded sources.
  \item \textbf{Evaluation under source exclusion, including probabilities.}
  We compare Lev-3B with decision models (\jev{}, \msdecision{}, \oaidecision{}), prompted frontier models, and a prompted
  instruction model on identical inputs, and evaluate raw and
  temperature-scaled probabilities, confident errors, and serving cost on
  represented and excluded sources.
  \item \textbf{Open release.} We release the checkpoint, adaptation recipe,
  data manifests, and prediction-level artifacts.
\end{enumerate}

The pointer architecture and temperature-fitting method follow \kev{}; the contribution here is the log-domain adaptation recipe, its evaluation under source exclusion, and the released artifacts. We do not claim that small language models for logs are new~\citep{lim2025logadreft,niloy2026multisource}.


\section{Related Work}

\paragraph{Log anomaly detection.}
Log anomaly detection is a long-standing problem in reliability engineering~\citep{he2016experience,he2021survey,landauer2023survey}.
Most pipelines first parse raw messages into templates~\citep{he2017drain,zhu2019tools} and then model the resulting events.
Early detectors used statistical features and mined invariants~\citep{xu2009detecting,lou2010invariants} or clustered log sequences~\citep{lin2016logcluster}.
Deep models then learned normal event sequences~\citep{du2017deeplog}, combined sequential and quantitative patterns~\citep{meng2019loganomaly}, used semantic event vectors to tolerate unstable logs~\citep{zhang2019logrobust}, reduced labeling cost through semi-supervision~\citep{yang2021plelog}, and removed the dependence on parsing~\citep{le2021neurallog}.
Transformer and language-model detectors followed~\citep{guo2021logbert,guan2024logllm}, alongside prompting-based approaches to log parsing and analysis~\citep{le2023chatgptparsing,qi2023loggpt,liu2024logprompt} and parameter-efficient adaptation across log corpora~\citep{lim2025logadreft}.
These methods typically target a single task and produce either a score or generated text.
We instead train one compact model for two log tasks that returns a probability vector over described options, and we study which training data produce that model.

\paragraph{Cross-source generalization.}
A detector evaluated on a held-out record from a familiar system faces a different problem from one evaluated on an unfamiliar logging system.
Event grouping, repeated patterns, and partition design can inflate scores on public corpora~\citep{landauer2024critical,le2022howfar}.
MetaLog studies generalization across logging systems~\citep{zhang2024metalog}, and LogICL retrieves cross-domain demonstrations, distinguishing target-domain few-shot adaptation from transfer without target access~\citep{ye2026logicl}. 
We adopt the stricter setting: HDFS and the withheld ATT\&CK sourcetypes enter no training, calibration, or model-selection step.
Unlike target-side methods such as demonstration retrieval, we improve transfer on the source side, by changing the training mixture.

\paragraph{Domain data, data mixtures, and synthetic data.}
Continued training on in-domain text improves downstream tasks in that domain~\citep{gururangan2020dont}, and LLM-generated examples are widely used to expand scarce supervision~\citep{wang2023selfinstruct}.
Skill-It shows that the order and mixture of training data matter: language models learn some skills faster when prerequisite skills are trained first, and a skill-aware data schedule can outperform uniform sampling~\citep{chen2023skillit}.
Training several tasks jointly can also cause negative transfer between them~\citep{yu2020gradient}.
We study these effects for log decisions.
Our recipe orders training from general natural-language decisions to log tasks, and our ablations add one data subset at a time to measure what each contributes.
Adding security supervision interferes with anomaly detection, and general log data, which is tied to neither evaluation task, offsets most of that interference and improves transfer to excluded sources.

\paragraph{Typed decision models.}
\jev{} exposes probabilities for binary, categorical, and ordinal questions~\citep{typesafe2026jev}.
\kev{} offers an open implementation and checkpoints with a compatible decision interface~\citep{palmer2026kev}, and Granite Decisions provides a Granite-based typed-decision implementation with head-training and calibration tools~\citep{granitedecisions2026}.
Hosted \jev{} has also been applied to log triage~\citep{jevlogs2026}, so applying a decision interface to logs is not by itself our contribution.
We contribute the log-domain data recipe, an open checkpoint trained with it, and a reliability evaluation under source exclusion that hosted endpoints do not provide.

\paragraph{Probability reliability.}
Proper scoring rules provide a principled way to assess predictive distributions~\citep{gneiting2007scoring}.
Temperature scaling adjusts confidence after training without changing the predicted class~\citep{guo2017calibration}.
Calibration measured on one distribution need not persist under shift~\citep{ovadia2019trust}.
In log anomaly detection, LoRD shows that confident errors can persist even when aggregate calibration metrics appear favorable~\citep{li2026lord}.
We therefore report proper scores, reliability diagrams, and confident false alarms and misses separately, and treat a single fitted temperature as a diagnostic rather than part of the method.

\paragraph{Small models for security classification.}
ATT\&CK technique identification has been evaluated with adapted small language models on multi-source cybersecurity logs~\citep{niloy2026multisource}.
Our task is constructed differently, from Splunk's attack-data repository~\citep{splunk2026attackdata}: a closed set of 130 parent techniques, with a held-out split of withheld sourcetypes.
It is neither a benign-versus-attack detector nor evidence of coverage of the entire ATT\&CK ontology (The full MITRE ATT\&CK catalog has 227 techniques and 479 sub-techniques).

\section{Model and Adaptation}
\label{sec:method}

Lev-3B is trained on four data sources, which we denote by a letter throughout the paper:
\begin{itemize}
  \item[\textbf{D}] \emph{general decision data}: bounded natural-language decision questions, containing no log data;
  \item[\textbf{A}] \emph{anomaly data}: window- and session-level anomaly questions from BGL, Spirit, and Thunderbird;
  \item[\textbf{T}] \emph{ATT\&CK data}: parent-technique classification questions from Splunk's attack-data repository, with synthetic augmentation of scarce techniques;
  \item[\textbf{L}] \emph{general log data}: auxiliary log decision questions from public benchmarks and generated or mechanically derived examples.
\end{itemize}
Training proceeds in two stages: stage one trains on D, and stage two continues from that checkpoint on a mixture of A, T, and L.
We write $X\rightarrow Y$ for training on $X$ followed by training on $Y$, and $Y+Z$ for a single stage on a mixture of $Y$ and $Z$.
The release checkpoint, Lev-3B, is D$\rightarrow$A+L+T, and Section~\ref{sec:exp-recipe} ablates each source.

\subsection{Decision interface}

An example is a tuple $(x,q,\mathcal{A},y)$, where $x$ is a log state, $q$ is a question, $\mathcal{A}=\{a_1,\ldots,a_K\}$ is an ordered set of described alternatives, and $y$ is the label.
For anomaly detection, the alternatives describe ordinary and anomalous behavior.
For security classification, they describe ATT\&CK parent techniques.
The model returns
\begin{equation}
  p_{\theta}(a_k\mid x,q,\mathcal{A}),\qquad
  \sum_{k=1}^{K}p_{\theta}(a_k\mid x,q,\mathcal{A})=1.
\end{equation}
The distribution is conditional on the supplied question and candidate set.
A probability for one technique is not a probability that the session is malicious, and adding or removing alternatives changes the decision problem.
The natural-language data in stage one and all three log sources in stage two use this format, so a single readout and loss serve every task.

\subsection{Backbone and probabilistic readout}

The backbone is Granite-4.1-3B-Base~\citep{granite2026}, trained and served with the open \kev{} implementation and checkpoint format~\citep{palmer2026kev}.
Following \kev{}, the state, question, and candidate descriptions are encoded with candidate-boundary markers and a final decision marker.
Let $h_k$ denote the hidden representation at the closing marker of candidate $k$ and $h_d$ the representation at the final decision marker.
A shared pointer readout computes
\begin{equation}
  z_k = \frac{(W_q h_d+b_q)^\top(W_{\mathrm{key}} h_k+b_{\mathrm{key}})}{\sqrt{d_p}},
  \qquad p_{\theta,k}=\softmax(z)_k,
  \label{eq:pointer}
\end{equation}
where $d_p$ is the projection dimension and both affine projections are shared across candidates.
The same readout therefore scores two anomaly alternatives in public log datasets or the full 130-technique catalog in ATT\&CK dataset~\citep{splunk2026attackdata}, without generating label text.
We train low-rank (LoRA) adapters and the pointer projections and keep the remaining backbone weights frozen.

Questions in the same request are isolated: each attends to the state and its own candidates, but not to another question's tokens or predicted answer.
Candidates within a question still interact causally, so option order is permuted during training.

\subsection{Stage one: natural-language decision adaptation}\label{sec:probs}
The first stage adapts the backbone and readout to bounded decisions expressed in natural language.
Training examples include binary propositions, categorical choices, and ordinal level selection, all in the format above, and contain no log data.
Appendix~\ref{app:decision-v7-details} describes the considered decision dataset, its source and question-type distributions, and the distinction between corpus counts and additional decision-only training exposure.
For a request $r$ containing $Q_r$ labeled questions, the loss first averages question losses within the request and then across requests:
\begin{equation}
  \mathcal{L}_{\mathrm{dec}}(\theta)
  =\E_r\left[-\frac{1}{Q_r}\sum_{q=1}^{Q_r}
       \sum_{k=1}^{K_{rq}}t_{rqk}\log p_{\theta,rqk}\right],
  \label{eq:loss}
\end{equation}
where $t_{rq}$ is a one-hot label or a supplied target distribution and $p_{\theta,rqk}$ is the corresponding option probability.
Options are permuted in every example.
\kev{}'s choice augmentation can also replace the correct option with ``none of the above,'' add an incorrect ``none'' option, or add a distractor; the rates used, and whether the 130-technique catalog (ATT\&CK dataset~\citep{splunk2026attackdata}) was kept intact during training.

Cross-entropy is a proper scoring rule~\citep{gneiting2007scoring}, but it does not guarantee calibration on a finite sample, under source shift, or after class balancing changes the effective training prevalence.
We therefore measure the learned probabilities 
rather than assume the objective has calibrated them. 

\subsection{Stage two: joint log adaptation}
\label{sec:stage2}
The second stage initializes from the stage-one adapter and pointer weights and trains on a mixture of three log data sources, each with the loss of Eq.~\eqref{eq:loss}:
\begin{equation}
  \mathcal{L}_{\mathrm{log}}
   =\lambda_{A}\mathcal{L}_{A}
    +\lambda_{T}\mathcal{L}_{T}
    +\lambda_{L}\mathcal{L}_{L},
   \qquad \lambda_{A}+\lambda_{T}+\lambda_{L}=1.
  \label{eq:mixture}
\end{equation}
The coefficients are the realized shares of each source in the request sampler, not separate loss weights.

\paragraph{Anomaly detection (A).}
Window- and session-level questions from BGL, Spirit, and Thunderbird~\citep{zhu2023loghub}, built as described in Section~\ref{sec:data}. 
Each question offers two described alternatives, ordinary and anomalous.

\paragraph{ATT\&CK classification with synthetic augmentation (T).}
We use records from Splunk's attack-data repository~\citep{splunk2026attackdata}, with the 130 parent techniques as candidate labels. The real training pool is small and long-tailed, containing 544 unique records. We synthesize examples for techniques below a target per-technique count:

\begin{itemize}
    \item \textbf{Select templates:} For each underrepresented technique, provide GPT-5.5 with one or two real training records for that technique. The records serve as templates for the sourcetype and field syntax.
    \item \textbf{Generate examples:} Instruct GPT-5.5 to create new, concrete log content---such as process names, paths, addresses, timestamps, and command lines---rather than paraphrase the templates.
    \item \textbf{Audit candidates:} Use a second GPT-5.5 pass to check technique plausibility and log-format correctness. Discard candidates that nearly duplicate an accepted example.
    \item \textbf{Prevent data leakage:} Draw templates only from training records; exclude development, calibration, and evaluation records, as well as records from any withheld sourcetype.
\end{itemize}
Real and synthetic support are reported separately throughout, and a technique with only
synthetic training examples is not called unseen.

\paragraph{General log data (L).}
General log data contains 28,740 auxiliary decision requests drawn from public log benchmarks and GPT-5.5 generated examples, spanning 35 task definitions (Appendix~\ref{app:general-log-details}). Its purpose is to expose the model to a wider variety of log formats and questions than the primary anomaly and ATT\&CK tasks provide, so that what it learns carries over to logging systems it has not seen.
No HDFS record or withheld-sourcetype record is used as a seed or template, and no generated question targets HDFS.

\paragraph{Training mixture.}
The training mixture contains 7,566 anomaly-detection examples (1,261 anomalous and 1,261 normal examples from each represented source), 28,740 general-log examples, and 3,945 ATT\&CK examples (544 real and 771 synthetic examples, repeated three times), totaling 40,251 training examples.

\paragraph{Release checkpoint.}
A single frozen checkpoint trained on D$\rightarrow$A+L+T produces every headline anomaly and technique result.
Ablated variants share the same stage-one checkpoint and drop one or more stage-two sources; the anomaly-only variant (D$\rightarrow$A) serves as the reference for interference from adding security supervision.

\subsection{Post-hoc temperature}
\label{sec:temperature}

Raw probabilities at inference temperature $T=1$ are our primary measure of the trained model.
As a separate diagnostic, we follow \kev{}'s release fitter and fit one scalar temperature, shared across both tasks, by minimizing question-level negative log-likelihood (NLL) over a fixed grid:
\begin{equation}
  p^{(T)}_k=\softmax(z/T)_k,
  \qquad
  T^*=\argmin_{T\in\mathcal G}
       \left[-\frac{1}{N_{\mathrm{cal}}}\sum_{i\in\mathcal{D}_{\mathrm{cal}}}
          \log p^{(T)}_{i,y_i}\right],
  \label{eq:temperature}
\end{equation}
where $\mathcal G=\{\exp(\log 0.25+(j/120)\log16):j=0,\ldots,120\}$ contains 121 log-spaced values between $0.25$ and $4$.
The calibration pool $\mathcal{D}_{\mathrm{cal}}$ is disjoint from training, development, and evaluation data at the parent-group level and contains no HDFS or withheld-sourcetype data, and no class priors derived from them.
The fitted value is applied unchanged to every source and both tasks.

A positive temperature preserves the predicted option and, for binary tasks, the ranking across examples and the decision at $0.5$.
It therefore cannot change accuracy or AUROC; it changes only the probabilities and decisions at other thresholds.
All probability scores are computed from full, unrounded probability vectors or saved logits, not from \kev{}'s derived confidence field.

\section{Data and Evaluation Setup}
\label{sec:data}
We evaluate Lev-3B on two bounded decision tasks: binary anomaly detection over log windows and sessions (Task~1), and closed-set classification over 130 ATT\&CK parent techniques (Task~2). Each task has \emph{represented sources}, whose training splits enter adaptation, and \emph{excluded sources}, which enter no project-controlled adaptation, calibration, or model-selection step. Excluded sources test transfer to logging systems unseen during adaptation. They are not a general out-of-distribution guarantee (Section~\ref{sec:limitations}).

\subsection{Task 1: Anomaly detection}
\label{sec:setup-anomaly}

\paragraph{Sources.} The adaptation mixture contains Blue Gene/L (BGL), Spirit, and Thunderbird, drawn from Loghub~\citep{zhu2023loghub}.
In Blue Gene/L (BGL), Spirit, and Thunderbird data, sessions are node-specific, timestamp-ordered groups of at most 100 log messages, with a shorter final group retained for each node when needed.
Each session is labeled anomalous if at least one constituent message has an anomaly label in the source data; otherwise, it is labeled normal. Session duration varies with the timestamps of its constituent messages.
Hadoop Distributed File System (HDFS) is the excluded source and is evaluated at block-session level.  Dataset release and grouping are recorded, since artifacts sharing a dataset name are not interchangeable~\citep{landauer2024critical}.

\paragraph{Exclusion.} HDFS is removed from all supervised tasks, auxiliary corpora, replay streams, and derived views in our pipeline. An HDFS template-classification example would violate this exclusion even without an anomaly label. This does not establish that HDFS is absent from the base model's pretraining (Section~\ref{sec:limitations}).

\paragraph{Partitioning.} Partitions are assigned before task construction. We check exact and normalized duplicates across partitions and group related incidents where identifiers permit. A split by node alone can leave a cluster-wide HPC event on both sides, and a split by block alone can leave related HDFS jobs on both sides. We report the grouping achievable with the source metadata.

\subsection{Task 2: ATT\&CK technique classification}
\label{sec:setup-attack}

\paragraph{Task.} The dataset comprises records derived from Splunk's attack-data repository~\citep{splunk2026attackdata}. Each question offers 130 parent techniques with fixed names and descriptions, under a pinned ontology revision and sub-technique-to-parent mapping. Uniform random choice has accuracy $1/130 \approx 0.77\%$, a reference for catalog size and not a competitive baseline under label imbalance. Model-visible state excludes repository paths, technique-directory names, and annotation headers that reveal the answer. Recordings with multiple technique annotations follow a single-label selection rule. 

\paragraph{Splits.}
The withheld split contains 178 records from 18 sourcetypes excluded from the ATT\&CK training, development, and calibration partitions. The withheld sourcetypes are selected through a seeded search targeting approximately 20\% of records while matching the overall file-size distribution.
Real records of the remaining sourcetypes are split per technique into train, development, calibration, and test sets, with at least one training record guaranteed before the other splits are filled.
We group files from the same attack execution where provenance permits.
Real and synthetic training support are recorded separately.

\subsection{Source roles}
\label{sec:setup-roles}

Source data are assigned to four roles, separated by parent group: parameter training, development (model, mixture, and threshold selection), probability calibration (fitting the single checkpoint-level temperature), and final evaluation. HDFS and the withheld sourcetypes are excluded from all three fitting roles and contribute only to final evaluation. Table~\ref{tab:data} gives the counts. 
Appendix~\ref{app:data-details} details the provenance, source composition, question types, and label distributions of general decision data and the general log data.
Prior target access is recorded in a ledger (Appendix~\ref{app:artifacts}).

\begin{table}[t]
\centering
\caption{Data and source access. Zeros are required exclusions. Training counts show unique records (real records only for ATT\&CK).}
\label{tab:data}
\begin{tabular}{llrrrr}
\toprule
Source and task & Unit & Train & Dev. & Cal. & Test \\
\midrule
BGL anomaly & Window/session & 55,775 & 9,295 & 9,297 & 18,592 \\
Spirit anomaly & Window/session & 34,624 & 3,808 & 3,977 & 7,858 \\
Thunderbird anomaly & Window/session & 61,758 & 10,294 & 10,293 & 20,587 \\
\textbf{HDFS anomaly (excluded)} & Block session & 0 & 0 & 0 & 115,013 \\
ATT\&CK, represented & Grouped record & 544 real & 51 & 38 & 134 \\
\textbf{ATT\&CK, withheld (excluded)} & Grouped record & 0 & 0 & 0 & 178 \\
General log data & \begin{tabular}[t]{@{}l@{}}Task-dependent\\(e.g., log line,\\log window, trace)\end{tabular} & 28,740 & -- & -- & -- \\
\bottomrule
\end{tabular}
\end{table}

\subsection{Evaluation Metrics}
\label{sec:setup-metrics}

\paragraph{Anomaly detection.}
We report AUROC and AUPRC (non-interpolated average precision) for ranking, and anomaly-class F1 at a fixed threshold of $0.5$ for every system.
A threshold selected on represented-source development data does not transfer to HDFS, and neither do thresholds aligned to the unlabeled HDFS score distribution (Appendix~\ref{app:threshold-transfer}), so we do not tune one.
AUPRC and F1 depend on prevalence and are compared only on matching populations.

\paragraph{ATT\&CK.} We report accuracy and macro-F1. Macro-F1 averages over the gold classes present in each frozen split, identical for every model, with zero-division fixed to zero. Full 130-class support and confusion counts are released so that other aggregations can be recomputed.

\paragraph{Probability quality.}
We report negative log-likelihood (NLL) from full probability vectors, which reflects both calibration and discrimination, and expected calibration error (ECE) over ten equal-width bins as a descriptive, bin-dependent diagnostic, with reliability diagrams.
For anomalies, bins use the predicted anomaly probability; for ATT\&CK, the top-label confidence.
We also report errors among predictions with confidence at least $0.9$, with the number accepted and coverage, split into confident false alarms and confident misses for anomalies.
Raw probabilities are taken at $T=1$; the fitted temperature is described in Section~\ref{sec:temperature}.

\section{Experiments}
\label{sec:experiments}

This section evaluates \method{} on the two tasks defined in Section~\ref{sec:data}. We organize it around three research questions.

\begin{description}
  \item[RQ1: Competitiveness and training recipe.] How does \method{} compare with hosted decision models (\jev{}, \msdecision{}, \oaidecision{}) and prompted frontier models, does the decision step followed by log adaptation outperform training on log data alone, and how do the data subsets affect results? (Sections~\ref{sec:compare}--\ref{sec:exp-recipe}.)
  \item[RQ2: General data and transfer.] Does adding general log data to task-specific adaptation improve performance on sources excluded from adaptation? (Section~\ref{sec:exp-transfer}.)
  \item[RQ3: Probabilities.] Are the predicted probabilities informative, and does the model make confident mistakes? (Sections~\ref{sec:probability} and~\ref{sec:exp-confidence}.)
\end{description}

We also compare inference latency and cost as a supporting analysis (Section~\ref{sec:latency-cost}). 

\subsection{Systems and evaluation setup}
\label{sec:exp-systems}
\paragraph{Systems.}
All systems receive the same visible log state, task wording, candidate descriptions, option order, and context limit; prompts and decoding rules are included in the release.
Data sources D, A, T, and L are defined in Section~\ref{sec:method}.
We train five decision-model variants, each a single run, and also evaluate the untrained backbone:
\begin{itemize}
  \item \textbf{Lev-3B (D$\rightarrow$A+L+T)}: the full recipe.
  \item \textbf{D$\rightarrow$A+T}: Lev-3B without general log data.
  \item \textbf{D$\rightarrow$A}: anomaly data only in stage two.
  \item \textbf{D}: stage one only; it has seen no log data.
  \item \textbf{A+L+T}: all log data in a single stage, without D. 
  \item \textbf{Granite-4.1-3B-Base}: the pretrained backbone with no adaptation.
\end{itemize}
We compare against a prompted instruction model, three hosted decision models (\jev{}, \msdecision{}, and \oaidecision{}), and two prompted frontier models:
\begin{itemize}
  \item \textbf{Granite-3B instruct}, the instruction-tuned backbone prompted zero-shot over the same options, without our training.
  \item \textbf{\jev{}}, a version-pinned hosted endpoint (1.13.0), evaluated zero-shot and 5-shot, with native probabilities evaluated as returned.
  \item \textbf{\msdecision{}}, a native decision endpoint evaluated zero-shot and 5-shot on anomaly detection and ATT\&CK, using the same ATT\&CK training demonstrations as Jev and GPT-5.6.
  \item \textbf{\oaidecision{}}, evaluated zero-shot and 5-shot on anomaly detection and ATT\&CK through the native Decisions API. Binary predictions use a $0.5$ probability threshold; ATT\&CK scores use the API-returned choice, including when returned probabilities tie. 
  \item \textbf{GPT-5.6 Luna and Sol}, prompted frontier models evaluated zero-shot and 5-shot. They return labels but not probability vectors, so they are not scored on probability quality.
\end{itemize}

\paragraph{Training settings.}
We use the same optimizer hyperparameters across the log-adaptation variants and do not perform hyperparameter tuning. Decision adaptation and joint log adaptation use effective batch size 8, LoRA rank 16, pointer projection dimension 256, and weight decay $0.01$. Frozen backbone weights use BF16, and training uses BF16 mixed precision while retaining LoRA adapters and the pointer head in FP32, with gradient checkpointing enabled.

Decision adaptation trains on the Decision-v7 subset (more details in Appendix~\ref{app:decision-v7-details}) of the decision data (D) for two epochs with learning rate $10^{-4}$ and seed 2. It then continues for one epoch on 1,425 additional examples from \kev{} covering date-based policy decisions and reasoning with missing evidence, mixed with 2,000 replayed Decision-v7 training examples, using learning rate $4\times10^{-5}$ and seed 1.  State, branch, and packed-token limits are 384, 1,024, and 2,048, respectively.

Joint log adaptation trains for one epoch with learning rate $2\times10^{-5}$ and seed 1. The full training mixture completes 5,032 optimizer updates. State, branch, and packed-token limits are 32,768, 65,536, and 65,536, respectively.

\paragraph{Evaluation conditions.} Each task is evaluated on represented sources (BGL, Spirit, Thunderbird; seen ATT\&CK sourcetypes) and excluded sources (HDFS; 18 withheld ATT\&CK sourcetypes).For \method{}, we evaluate raw probability vectors at $T=1$ and probabilities after the single fitted temperature. For \jev{}, \msdecision{}, and \oaidecision{}, we evaluate native probabilities as returned, without additional temperature fitting. 
\label{sec:exp-compare}

\begin{table}[H]
\centering\small
\caption{Anomaly detection: comparison with prompted and hosted systems. TB denotes Thunderbird. F1 is anomaly-class F1 in percent; AUROC and AUPRC are on $[0,1]$. HDFS is excluded from adaptation.  \textbf{Bold}: best in column.  Dashes denote unavailable continuous scores. $^*$ denote metrics computed on a 10\% sample of the data.}
\label{tab:anomaly-main}
\begin{tabularx}{\textwidth}{@{}Y r r r r r r@{}}
\toprule
 & \multicolumn{3}{c}{Represented (F1)} & \multicolumn{3}{c}{HDFS (excluded)} \\
\cmidrule(lr){2-4}\cmidrule(lr){5-7}
System & BGL & Spirit & TB & AUROC & AUPRC & F1 \\
\midrule
Granite-4.1-3B-Base & 0.19 & 0.00 & 0.44 & 0.6063 & 0.1579 & 12.21 \\
Granite-3B instruct, prompted & 63.59 & 16.36 & 2.41 & -- & -- & 14.93 \\
\jev{} (0-shot) & 57.67 & 18.12 & 3.12 & 0.6168 & 0.0537 & 5.65 \\
\jev{} (5-shot) & 57.68 & 21.16 & 6.55 & 0.7074 & 0.0958 & 6.54 \\
\msdecision{} (0-shot) & 60.03 & 21.98 & 3.10 & 0.7162 & 0.2376 & 14.29 \\
\msdecision{} (5-shot) & 61.35 & 22.14 & 4.84 & 0.7533 & 0.2408 & 15.07 \\
\oaidecision{} (0-shot) & 59.07$^*$ & 18.52$^*$ & 2.45$^*$ & \textbf{0.7967}$^*$ & 0.2072$^*$ & 25.91$^*$ \\
\oaidecision{} (5-shot) & 71.80$^*$ & 28.25$^*$ & 5.21$^*$ & 0.7861$^*$ & 0.3233$^*$ & 20.90$^*$ \\
GPT-5.6 Luna (0-shot) & 58.71 & 16.25 & 3.75 & -- & -- & 21.92 \\
GPT-5.6 Luna (5-shot) & 68.64$^*$ & 22.83$^*$ & 4.86$^*$ & -- & -- & 15.77$^*$ \\
GPT-5.6 Sol (0-shot) & 58.17 & 19.32 & 4.33 & -- & -- & 17.05 \\
GPT-5.6 Sol (5-shot) & 71.73$^*$ & 26.07$^*$ & 9.68$^*$ & -- & -- & 14.23$^*$ \\
\midrule
\rowcolor{ours} \method{} & \textbf{99.36} & \textbf{88.33} & \textbf{88.20} & 0.7519 & \textbf{0.4116} & \textbf{43.32} \\
\bottomrule
\end{tabularx}
\end{table}

\subsection{Comparison with hosted decision models and frontier models}\label{sec:compare}

In this experiment, we ask whether a small adapted model can compete with hosted models that are prompted but not trained on our sources. We compare \method{} with Granite-4.1-3B-Base, prompted Granite-3B instruct, \jev{}, \msdecision{}, \oaidecision{}, and GPT-5.6 Luna and Sol on identical inputs, with \jev{}, \msdecision{}, \oaidecision{} and GPT-5.6 tested zero-shot and 5-shot. We report F1, AUROC and AUPRC for anomaly detection and accuracy and macro-F1 for ATT\&CK (Tables~\ref{tab:anomaly-main} and~\ref{tab:attack-main}). 

\textbf{Adaptation matters most on logs.} On the sources \method{} was adapted on, the gap is large. Prompted models achieve at most about 28\% and 10\% F1 on Spirit and Thunderbird, even with five demonstrations, while \method{} achieves more than 85\% on both. Instruction tuning alone does not close the gap: prompted Granite-3B instruct is comparable to the weakest hosted systems on Spirit and Thunderbird and far below the hosted systems on ATT\&CK, although it is competitive on BGL ($63.59$ F1). The advantage comes from adaptation and not from the backbone.

\textbf{Out-of-distribution anomaly detection is hard for everyone, but \method{} is still ahead.} On HDFS, \method{}'s F1 drops sharply relative to the HPC sources, and its AUROC is only moderate, so transfer to an unseen logging system is far from solved. All other systems are also low on HDFS, and \method{} has the best F1 and AUPRC of all systems, while \msdecision{} and \oaidecision{} has higher HDFS AUROC
Demonstrations do not help consistently. \jev{} improves with five examples on HDFS (F1 $5.65\to6.54$, AUROC $0.6168\to0.7074$, AUPRC $0.0537\to0.0958$). 

\textbf{On ATT\&CK, \method{} beats all three decision models (\jev{}, \msdecision{} and \oaidecision{}) but not the frontier models.} A 3B model trained on a few hundred real records outperforms \jev{}, \msdecision{} and \oaidecision{} on every ATT\&CK column, but prompted GPT-5.6 is better, especially on withheld sourcetypes. A plausible reason is that the frontier models bring broad prior knowledge of ATT\&CK techniques, whereas \method{} can only learn techniques it saw. The training data are also sparse: the pool contains only 544 unique real records across 130 parent techniques, leaving many techniques with limited real examples.

\begin{table}[H]
\centering\small
\caption{ATT\&CK parent-technique classification with 130 offered alternatives: comparison with prompted and hosted systems. All metrics are percentages. 
Every row uses identical inputs and options. Withheld sourcetypes contribute no training data, and some withheld-split techniques have no real training records. \colorbox{ours}{Shaded}: our model. \textbf{Bold}: best in column. \underline{Underline}: best among non-frontier systems.
}
\label{tab:attack-main}
\begin{tabularx}{\textwidth}{@{}Y r r r r@{}}
\toprule
 & \multicolumn{2}{c}{Represented sourcetypes} & \multicolumn{2}{c}{Withheld sourcetypes} \\
\cmidrule(lr){2-3}\cmidrule(lr){4-5}
System & Accuracy & Macro-F1 & Accuracy & Macro-F1 \\
\midrule
Granite-4.1-3B-Base & 0.00 & 0.00 & 0.00 & 0.00 \\
Granite-3B instruct, prompted & 3.0 & 0.1 & 5.1 & 1.6 \\
\jev{} (0-shot) & 24.6 & 20.6 & 32.6 & 28.1 \\
\jev{} (5-shot) & 29.9 & 23.2 & 34.3 & 25.6 \\
\msdecision{} (0-shot) & 22.4 & 17.3 & 28.1 & 23.9 \\
\msdecision{} (5-shot) & 25.4 & 19.5 & 24.7 & 20.3 \\
\oaidecision{} (0-shot) & 28.4 & 24.1 & 38.2 & 28.1 \\
\oaidecision{} (5-shot) & 32.8 & 25.2 & 41.6 & 26.6 \\
GPT-5.6 Luna (0-shot) & 38.8 & 31.5 & 51.7 & 42.7 \\
GPT-5.6 Luna (5-shot) & 38.1 & 30.2 & 53.9 & 38.6 \\
GPT-5.6 Sol (0-shot) & 42.5 & \textbf{36.9} & 51.7 & 45.4 \\
GPT-5.6 Sol (5-shot) & \textbf{45.5} & 36.1 & \textbf{56.2} & \textbf{49.4} \\
\midrule
\rowcolor{ours} \method{} & \underline{35.07} & \underline{30.73} & \underline{42.13} & \underline{29.53} \\
\bottomrule
\end{tabularx}
\end{table}


\takeaway{Adaptation, not prompting, drives performance on logs: \method{} far outperforms hosted models on the sources it was adapted on and remains ahead on an out-of-distribution (unseen) log source, though with much lower absolute performance. On security classification it beats decision models (\jev{}, \msdecision{}, \oaidecision{}) but trails the frontier models.}

\subsection{Comparison with prior log anomaly detectors}
\label{sec:exp-prior}

Tables~\ref{tab:anomaly-main} and~\ref{tab:attack-main} compare \method{} with systems evaluated on identical inputs. Here, we compare Lev with published detectors for cross-system log anomaly detection.
We follow the multi-source protocol of LogICL~\citep{ye2026logicl}: a model is trained on two HPC sources, a primary and an auxiliary one, and tested on a third logging system.
Because this protocol fixes its own training sources, we train dedicated Lev variants rather than evaluating the release checkpoint.
Each variant undergoes log adaptation on only anomaly data from the two listed sources only, without T or L.
These variants therefore measure how well the decision interface and backbone transfer across systems, not the effect of the data recipe, which Sections~\ref{sec:exp-recipe} and~\ref{sec:exp-transfer} isolate.
Baselines are as reported by \citep{ye2026logicl}, and Lev uses the same source pairs and target systems; per-source budgets are in the release manifest.

\begin{table}[H]
\centering\small
\caption{Comparison with prior cross-system log anomaly detectors under the multi-source protocol of LogICL. Each cell is precision / recall / F1 (percent) for training on the two listed sources (primary first) and testing on the target. Lev rows are variants with anomaly-only stage-two training, not the release checkpoint. LogPrompt is prompted zero-shot and uses no training data. Baseline numbers are from \citep{ye2026logicl}. Lib: Liberty; TB: Thunderbird. Bold: best F1 per column.}

\label{tab:transfer}
{\scriptsize
\setlength{\tabcolsep}{2pt}
\begin{tabularx}{\textwidth}{
  @{}
  >{\raggedright\arraybackslash}p{2.75cm}
  *{5}{>{\centering\arraybackslash}X}
  @{}
}
\toprule
\textbf{Model}
  & \shortstack{BGL+Lib\\$\to$ TB}
  & \shortstack{Lib+BGL\\$\to$ TB}
  & \shortstack{TB+Lib\\$\to$ BGL}
  & \shortstack{BGL+TB\\$\to$ Lib}
  & \shortstack{TB+Lib\\$\to$ HDFS} \\
\midrule
LogRobust~\citep{zhang2019logrobust}
  & 57.36/91.79/70.60 & 18.07/4.59/7.32 & 66.48/31.69/42.92 & 79.15/5.29/9.91 & 2.43/82.50/4.73 \\
LogPrompt~\citep{liu2024logprompt}
  & 51.10/70.65/59.30 & \textemdash & 39.49/76.72/52.15 & 66.54/47.30/55.29 & 8.73/84.07/15.81 \\
MetaLog~\citep{zhang2024metalog}
  & 42.10/86.23/56.58 & 56.70/73.50/64.01 & 29.16/23.35/25.93 & 90.72/10.02/18.04 & 3.50/0.05/0.10 \\
LogICL~\citep{ye2026logicl}
  & 70.16/79.83/74.68 & 65.96/66.06/66.01 & 40.21/34.71/37.25 & 82.39/69.70/75.52 & 72.87/22.75/\textbf{34.67} \\
\rowcolor{ours} Lev (two-source)
  & 83.21/97.88/\textbf{89.95} & 86.15/94.52/\textbf{90.14} & 85.21/47.40/\textbf{60.91} & 96.97/93.22/\textbf{95.06} & 100.00/19.86/33.14 \\
\bottomrule
\end{tabularx}
}
\end{table}

\paragraph{The decision model transfers across HPC systems better than prior detectors.}
Lev has the best F1 on all four HPC targets, 9--24 points above the strongest baseline on each.
It has the highest recall on three of the four HPC targets, and on BGL+TB$\to$Lib it reaches 93.22 recall against 69.70 for LogICL, while also having the highest precision on every target.
The hardest target is BGL (TB+Lib$\to$BGL), where recall falls to 47.40, although F1 still exceeds every baseline.

\paragraph{On HDFS, Lev matches LogICL with a conservative failure mode.}
On HDFS, Lev is close to LogICL (33.14 vs.\ 34.67 F1), with perfect precision but low recall: it flags few sessions, and every session it flags is anomalous.
This is the same pattern the release checkpoint shows on HDFS, where nearly all confident errors are missed anomalies (Section~\ref{sec:exp-confidence}).
It suggests that the tendency to miss anomalies on HDFS comes from transfer to a structurally different logging system, not from the multi-task mixture.

\subsection{Effect of training recipe}
\label{sec:exp-recipe}

In this experiment, we ask which parts of the recipe produce the results, by training variants of \method{} that differ in which data subsets they see and  how training is staged.  The variants add subsets cumulatively: decision data only (D), anomaly adaptation (D$\rightarrow$A), anomaly plus ATT\&CK (D$\rightarrow$A+T), and the full recipe (D$\rightarrow$A+L+T), which is \method{}. A log-data-only variant trains on A+L+T in a single stage without the decision step, which is meant to test the value of incremental skills as compared to teaching the model to decide and adapt to logs simultaneously, and the untrained base model gives the reference point. All variants are evaluated on every source (Tables~\ref{tab:anomaly-ablation} and~\ref{tab:attack-ablation}). Each adjacent pair of cumulative variants differs by one subset, which makes the comparisons controlled in design; each is still a single run, so the differences carry no variance estimate.

\textbf{Task-specific data drives task performance; generic decision training alone is not enough.}
The untrained base model scores near zero F1 on the HPC anomaly sources (below $0.5$). The decision step alone lifts BGL F1 to $65.93$ and Spirit to $17.97$, but leaves Thunderbird near zero ($2.82$) and ATT\&CK at a few percent accuracy (uniform chance over 130 techniques is $0.77\%$). Anomaly adaptation then turns it into a strong HPC detector but leaves ATT\&CK at the decision-only level, so strong skill on a task appears only when that task's data is in the mix, and anomaly data does not help ATT\&CK. On HDFS, the decision step raises AUROC ($0.6063\to0.6981$) and AUPRC ($0.1579\to0.2724$) but lowers F1 ($12.21\to9.05$), so its effect there is mixed.

\textbf{Adding ATT\&CK data enables Task~2, at a large cost to anomaly detection.} Going from D$\rightarrow$A to D$\rightarrow$A+T raises represented ATT\&CK accuracy from $4.48$ to $35.8$ and macro-F1 from $4.02$ to $31.7$, the largest gain in the ablation. The anomaly cost is also large: Spirit F1 falls from $91.82$ to $77.47$, Thunderbird from $94.63$ to $91.36$, and on HDFS AUROC falls from $0.7787$ to $0.7509$, AUPRC from $0.3117$ to $0.2485$ and F1 from $31.23$ to $25.31$. BGL is essentially unchanged ($99.64\to99.30$).

\textbf{Adding general log data recovers part of that cost and helps on unseen sources.} Going from D$\rightarrow$A+T to the full recipe raises withheld ATT\&CK accuracy from $34.3$ to $42.1$ and macro-F1 from $19.8$ to $29.5$, while the represented sourcetypes change by about a point (slightly lower).  On anomaly detection, L recovers Spirit F1 ($77.47\to88.33$) and substantially raises HDFS AUPRC ($0.2485\to0.4116$) and F1 ($25.31\to43.32$), with HDFS AUROC nearly unchanged ($0.7509\to0.7519$) and BGL stable ($99.30\to99.36$). It does not help Thunderbird, which falls further ($91.36\to88.20$). This comparison is the basis for RQ2 (Section~\ref{sec:exp-transfer}).

\textbf{Anomaly detection pays a price for the extra tasks.} Relative to anomaly-only adaptation, the full recipe is lower on all three HPC sources (BGL $99.64\to99.36$, Spirit $91.82\to88.33$, Thunderbird $94.63\to88.20$) and on HDFS AUROC ($0.7787\to0.7519$), but higher on HDFS AUPRC and F1. The D$\rightarrow$A+T row shows that in-domain loss and the HDFS AUROC loss appear when T is added, and L offsets part of it, apart from Thunderbird. The extra subsets therefore appear to redistribute capacity, trading in-domain anomaly accuracy for ATT\&CK skill and, with L, HDFS precision.

\textbf{Starting from the decision checkpoint helps with transfer.}
\method{} and the log-data-only variant (A+L+T) see the same stage-two data and differ only in whether stage one runs first.
On represented sources the two perform comparably, so log data alone is enough to fit the sources seen in training.
The difference appears on excluded sources.
On HDFS, the log-data-only variant has lower AUPRC and F1 than even the untrained base model, whereas \method{} more than doubles both (AUPRC 0.10 vs.\ 0.41 and F1 11.8 vs.\ 43.3 for the log-data-only variant and \method{}, respectively).
On ATT\&CK, stage one helps on both splits but most on withheld sourcetypes (28.1 vs.\ 42.1 accuracy).

\begin{table}[H]
\centering\small
\caption{Anomaly detection: ablation of training data and stages. D: general decision data; A: anomaly data; L: general log data; T: ATT\&CK data. A check mark means the component is in the training data. Steps is the number of training stages: in the two-step recipe, D comes first and the remaining components follow; in log-data-only, the log data are trained in a single stage without D; the base model has no adaptation (0). \colorbox{ours}{Shaded}: our models. \textbf{Bold}: best in column among these rows. HDFS is excluded from adaptation.}
\label{tab:anomaly-ablation}
\begin{tabularx}{\textwidth}{@{}Y c c c c c r r r r r r@{}}
\toprule
 & \multicolumn{4}{c}{Training data} & & \multicolumn{3}{c}{Represented (F1)} & \multicolumn{3}{c}{HDFS (excluded)} \\
\cmidrule(lr){2-5}\cmidrule(lr){7-9}\cmidrule(lr){10-12}
Variant & D & A & L & T & Steps & BGL & Spirit & TB & AUROC & AUPRC & F1 \\
\midrule
 Granite-4.1-3B-Base (no adaptation) & & & & & 0 & 0.19 & 0.00 & 0.44 & 0.6063 & 0.1579 & 12.21 \\
\rowcolor{ours} Log data only (A+L+T) & & \checkmark & \checkmark & \checkmark & 1 & 99.33 & 85.96 & 92.71 & 0.7154 & 0.1039 & 11.80 \\
\rowcolor{ours} Granite-4.1-3B-Base $\rightarrow$ D & \checkmark & & & & 1 & 65.93 & 17.97 & 2.82 & 0.6981 & 0.2724 & 9.05 \\
\rowcolor{ours} D$\rightarrow$A & \checkmark & \checkmark & & & 2 & \textbf{99.64} & \textbf{91.82} & \textbf{94.63} & \textbf{0.7787} & 0.3117 & 31.23 \\
\rowcolor{ours} D$\rightarrow$ A+ T & \checkmark & \checkmark & & \checkmark & 2 & 99.30 & 77.47 & 91.36 & 0.7509 & 0.2485 & 25.31 \\
\rowcolor{ours} \method{} & \checkmark & \checkmark & \checkmark & \checkmark & 2 & \underline{99.36} & \underline{88.33} & \underline{88.20} &\underline{ 0.7519} & \textbf{0.4116} & \textbf{43.32} \\
\bottomrule
\end{tabularx}
\end{table}


\begin{table}[H]
\centering\small
\caption{ATT\&CK \textbf{}classification: ablation of training data and stages. D, A, L, T and Steps as in Table~\ref{tab:anomaly-ablation}. All metrics are percentages. \colorbox{ours}{Shaded}: our models. \textbf{Bold}: best in column among these rows.}
\label{tab:attack-ablation}
\begin{tabularx}{\textwidth}{@{}Y c c c c c r r r r@{}}
\toprule
 & \multicolumn{4}{c}{Training data} & & \multicolumn{2}{c}{Represented} & \multicolumn{2}{c}{Withheld} \\
\cmidrule(lr){2-5}\cmidrule(lr){7-8}\cmidrule(lr){9-10}
Variant & D & A & L & T & Steps & Acc. & Macro-F1 & Acc. & Macro-F1 \\
\midrule
 Granite-4.1-3B-Base (no adaptation) & & & & & 0 & 0.00 & 0.00 & 0.00 & 0.00 \\
\rowcolor{ours} Log data only (A+L+T) & & \checkmark & \checkmark & \checkmark & 1 & 29.10 & 24.27 & 28.09 & 24.34 \\
\rowcolor{ours} Granite-4.1-3B-Base $\rightarrow$ D & \checkmark & & & & 1 & 4.48 & 4.02 & 8.43 & 3.89 \\
\rowcolor{ours} D$\rightarrow$A & \checkmark & \checkmark & & & 2 & 4.48 & 4.02 & 5.06 & 4.89 \\
\rowcolor{ours} D$\rightarrow$A+T & \checkmark & \checkmark & & \checkmark & 2 & \textbf{35.8} & \textbf{31.7} & 34.3 & 19.8 \\
\rowcolor{ours} \method{}  & \checkmark & \checkmark & \checkmark & \checkmark & 2 & \underline{35.07} & \underline{30.73} & \textbf{42.13} & \textbf{29.53} \\
\bottomrule
\end{tabularx}
\end{table}


\takeaway{Task skill comes mainly from task data: the decision step alone provides little, and adding a task's data enables that task. Adding ATT\&CK data costs anomaly accuracy, and adding general log data recovers part of that cost and improves transfer to unseen sources, though not on every source.}

\subsection{RQ2: General data and transfer to excluded sources}
\label{sec:exp-transfer}

In this experiment, we ask whether general log data improves transfer to sources that enter no adaptation, calibration or model-selection step. The controlled comparison is \method{} against the D$\rightarrow$A+T (Anomaly+Attack) variant, which differs only in lacking L, on both HDFS and the withheld ATT\&CK sourcetypes (Tables~\ref{tab:anomaly-ablation} and~\ref{tab:attack-ablation}). HDFS is evaluated at block-session level on $115{,}013$ sessions with a fixed classification threshold of $0.5$, and the withheld split has $178$ records from $18$ sourcetypes. 

\textbf{On unseen ATT\&CK sourcetypes, general data helps.} \method{} is better than the variant without general log data on withheld sourcetypes in both accuracy ($34.3\to42.1$) and macro-F1 ($19.8\to29.5$), while giving up only a point or less on represented sourcetypes. The accuracy gain is about $14$ records of $178$ from a single run with no interval, so this is suggestive evidence for RQ2 and not a settled result. 

\textbf{On HDFS, general data improves precision-oriented measures and leaves ranking nearly unchanged.} Against the Anomaly+Attack variant (D$\rightarrow$A+T), \method{} has higher AUPRC ($0.2485\to0.4116$) and F1 ($25.31\to43.32$) and nearly identical AUROC ($0.7509\to0.7519$). In practice \method{} finds more of the true anomalies among its most suspicious sessions, which matters for alert triage. Against anomaly-only adaptation (D$\rightarrow$A), \method{} has higher AUPRC and F1 but lower AUROC ($0.7787\to0.7519$) and lower F1 on the HPC sources; because that comparison changes both L and T, and the AUROC drop already appears when T alone is added, it should not be attributed to general log data.

\takeaway{General log data improves transfer to unseen sources at almost no in-domain cost on ATT\&CK, and improves HDFS AUPRC and F1 with little change in AUROC.}

\subsection{Raw and calibrated probability quality}
\label{sec:probability}

In this experiment, we ask whether \method{}'s probabilities are informative, and how much a single post-hoc temperature changes them. We evaluate \emph{raw} probabilities at $T=1$ and \emph{fitted} probabilities, using one scalar temperature that minimizes NLL on an independent calibration pool and is applied unchanged to every source and both tasks. A positive temperature preserves the argmax and, for binary tasks, the ranking, so accuracy, F1 at $0.5$ and AUROC do not change. We report NLL, which combines calibration and discrimination, and ECE as a descriptive, bin-dependent diagnostic (Table~\ref{tab:calibration});  Native \jev{}, \msdecision{} and \oaidecision{} probabilities are evaluated as returned, and the frontier models return no probability vectors.

\textbf{Source-prior baselines.} The constant anomaly prior is the anomalous fraction among the pooled $23{,}397$ source-development records from BGL, Spirit, and Thunderbird: $4{,}177/23{,}397=0.178527$. Records are pooled by count without upsampling or balancing datasets. The ATT\&CK class prior uses the $51$ represented-source development records and fixed add-one smoothing over the $130$ offered techniques, $q_k=(n_k+1)/(51+130)$. Each baseline returns the same distribution for every target record. Classification ties use the first ID in the fixed sorted catalog. Neither HDFS nor withheld-sourcetype labels are used to estimate these distributions or choose the smoothing rule; target labels are used only for evaluation.

\begin{table}[t]
\centering\small
\caption{Probability quality before (raw) and after a single checkpoint-level temperature fit (fit). Lower is better. Uniform choice over 130 techniques has NLL $=\ln 130\approx4.87$. \colorbox{ours}{Shaded rows}: our model. \textbf{Bold}: best per source and metric among rows shown. \jev{} is native only; prior baselines use no excluded-source labels.$^*$ marks metrics computed on a 10\% sample of the corresponding test set. }
\label{tab:calibration}
\begin{tabular}{@{}l r r r r@{}}
\toprule
& \multicolumn{2}{c}{NLL} & \multicolumn{2}{c}{ECE} \\
\cmidrule(lr){2-3}\cmidrule(lr){4-5}
Source / system & Raw & Fit & Raw & Fit \\
\midrule
\multicolumn{5}{@{}l}{\emph{Represented}} \\
\rowcolor{ours} BGL anomaly & 0.0189 & \textbf{0.0173} & \textbf{0.0046} & 0.0070 \\
\rowcolor{ours} Spirit anomaly & 0.0791 & \textbf{0.0578} & 0.0094 & \textbf{0.0025} \\
\rowcolor{ours} Thunderbird anomaly & \textbf{0.0259} & 0.0286 & \textbf{0.0083} & 0.0166 \\
\rowcolor{ours} ATT\&CK represented & 3.8512 & \textbf{3.1198} & 0.3099 & \textbf{0.1272} \\
\midrule
\multicolumn{5}{@{}l}{\emph{Excluded}} \\
\rowcolor{ours} HDFS anomaly, \method{} & 0.2050 & \textbf{0.1236} & 0.0212 & \textbf{0.0186} \\
HDFS, \jev{} (0-shot) & 1.2179 & -- & 0.6114 & -- \\
HDFS, \jev{} (5-shot) & 0.9207 & -- & 0.5160 & -- \\
HDFS, \msdecision{} (0-shot) & 0.4128 & -- & 0.2219 & -- \\
HDFS, \msdecision{} (5-shot) & 0.3714 & -- & 0.1821 & -- \\
HDFS, \oaidecision{} (0-shot)$^*$ & 0.2528 & -- & 0.0983 & -- \\
HDFS, \oaidecision{} (5-shot)$^*$ & 0.2953 & -- & 0.1418 & -- \\
HDFS, constant anomaly prior & 0.2414 & -- & 0.1492 & -- \\
\rowcolor{ours} ATT\&CK withheld, \method{} & 2.7903 & \textbf{2.4989} & 0.2751 & 0.1441 \\
ATT\&CK withheld, \jev{} (0-shot) & 5.7263 & -- & 0.2799 & -- \\
ATT\&CK withheld, \jev{} (5-shot) & 6.0517 & -- & 0.2556 & -- \\
ATT\&CK withheld, \msdecision{} (0-shot) & 3.9519 & -- & 0.3810 & -- \\
ATT\&CK withheld, \msdecision{} (5-shot) & 4.2513 & -- & 0.4237 & -- \\
ATT\&CK withheld, \oaidecision{} (0-shot) & 6.3435 & -- & 0.1308 & -- \\
ATT\&CK withheld, \oaidecision{} (5-shot) & 4.9120 & -- & 0.1176 & -- \\
ATT\&CK withheld, smoothed class prior & 4.5593 & -- & \textbf{0.0172} & -- \\
\bottomrule
\end{tabular}
\end{table}

\textbf{On anomaly detection, probability quality degrades on the excluded source.} Raw NLL is lowest on the sources \method{} was adapted on and higher on HDFS: about $11\times$ BGL, $8\times$ Thunderbird and $2.6\times$ Spirit. On HDFS, raw \method{} has lower reported NLL ($0.2050$) and ECE ($0.0212$) than the native probabilities of all three hosted decision models.

\begin{figure}
\centering
\includegraphics[
  width=0.9\textwidth
]{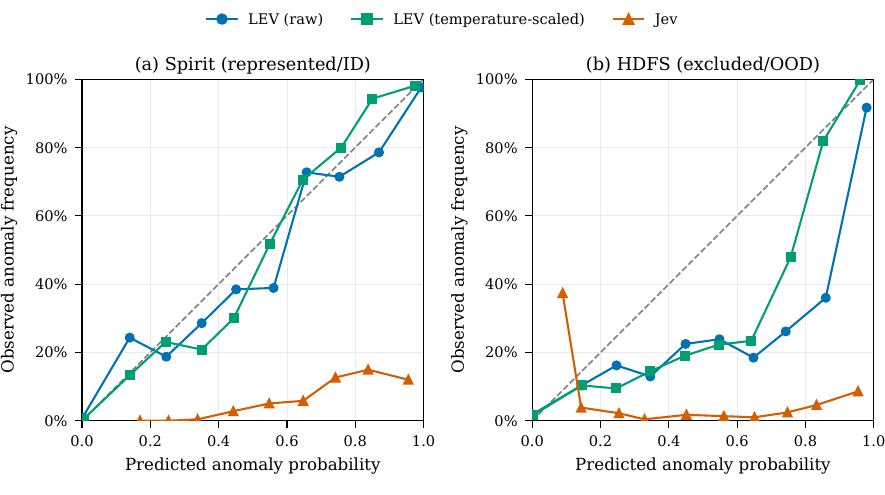}
\caption{Anomaly-probability reliability on (a) Spirit, a represented source, and (b) HDFS, which is excluded from adaptation. The dashed diagonal is perfect calibration. Curves show raw \method{}, fitted-temperature \method{} and native \jev{} probabilities in ten equal-width bins.}

\label{fig:reliability}
\end{figure}

\textbf{Reliability diagrams show where the probabilities go wrong.} Figure~\ref{fig:reliability} complements Table~\ref{tab:calibration}. We group sessions into ten equal-width bins by predicted anomaly probability and plot the fraction of truly anomalous sessions in each bin. A perfectly calibrated model lies on the dashed diagonal; a curve below it overstates the chance of an anomaly, and a curve above it understates it. We compare raw \method{}, fitted \method{} and native \jev{} on two sources chosen for contrast: Spirit, which \method{} was adapted on, and HDFS, which it never saw.

On Spirit (a), both \method{} curves stay close to the diagonal, with some scatter; we observe similar results for other in-domain datasets (BGL and Thunderbird). On HDFS (b), raw \method{} falls well below the diagonal in the upper bins: sessions given a high anomaly probability are anomalous much less often than predicted, and only the top bin approaches the diagonal. The fitted temperature moves the upper-bin points closer to the diagonal but does not remove the gap. \jev{}'s curve sits near zero observed frequency over almost the whole range on both sources, so its anomaly probabilities carry little information.


\textbf{ATT\&CK probabilities are only loosely informative.} Raw NLL scores are better than the uniform reference, so the probabilities carry some signal, but NLL stays high and ECE is large, even after fitting. \jev{}'s native ATT\&CK NLL is worse than the uniform reference, meaning its probabilities are confidently wrong on average. We therefore describe \method{}'s ATT\&CK probabilities as informative but not reliable. \oaidecision{}’s lower ECE reflects closer agreement between average selected-label confidence and accuracy within the ten confidence bins. NLL instead evaluates the probability assigned to the true technique. In the five-shot run, \oaidecision{}'s returned vectors assign zero probability to the true technique on 18 of 178 records, producing large NLL penalties under the declared probability floor. Thus lower ECE can coexist with worse NLL.

\takeaway{On anomaly detection, raw \method{} probabilities are well calibrated on adapted sources but less so on HDFS. Across both excluded sources, raw \method{} has lower reported NLL and ECE than all three hosted decision models, except for withheld ATT\&CK ECE, where \oaidecision{} has the lowest value among these models. ATT\&CK probabilities are informative but not reliable.}

\subsection{Confidence on errors}
\label{sec:exp-confidence}
In this experiment, we ask whether aggregate scores hide confident errors. A prediction is \emph{accepted} if its top-label confidence is at least $\tau=0.9$; \emph{coverage} is the fraction of examples accepted, and the \emph{confident error rate} is the fraction of accepted predictions that are wrong (Table~\ref{tab:errors}). For anomaly detection we separate confident false alarms (benign sessions with anomaly probability $\geq 0.9$) from confident misses (anomalous sessions with anomaly probability $\leq 0.1$); these are counts, not rates. 
The value $\tau=0.9$ is a fixed reporting point, not an operating threshold. Because coverage differs across systems, confident error rates are not directly comparable. We compare raw and fitted \method{} with native \jev{}, \msdecision{}, and \oaidecision{} on the excluded sources.

\textbf{\method{} and \jev{} fail in opposite directions on HDFS.} \jev{} is confident on only about $7\%$ of sessions, and most of those predictions are wrong, almost all of them false alarms. \method{} is confident on about $98\%$ of sessions, and only about $2\%$ of those are wrong, but nearly all of its confident errors are misses: $2{,}105$ anomalous sessions are called normal with high confidence, against $84$ confident false alarms. The error rates should not be compared directly, because the coverages differ by more than an order of magnitude and the benign sessions dominate \method{}'s accepted set. The practical difference is that \jev{}'s confidence cannot be used to filter its alerts, whereas \method{}'s confidence keeps the alert stream clean but hides missed anomalies.

\textbf{A single temperature reduces confident errors modestly and does not remove them.} On HDFS, the fitted temperature removes nearly all confident false alarms ($84\to1$) and about a tenth of the confident misses ($2{,}105\to1{,}912$), and confident coverage falls slightly ($98.3\%\to96.3\%$), consistent with softer probabilities. Roughly $1{,}900$ sessions are still missed with high confidence, so calibration alone does not address this failure mode.

\textbf{On unseen ATT\&CK sourcetypes, confidence is moderately informative.} Overall, \method{} is wrong on $57.9\%$ of the withheld records. Among the $70$ predictions with confidence at least $0.9$ (coverage $39.3\%$), $37.1\%$ are wrong. After fitting, the confident set shrinks to $18$ records, with $33.3\%$ wrong (coverage $10.1\%$). Confidence therefore concentrates correct predictions, but about a third of the most confident predictions are still wrong, so it is not a reliable filter. Native \jev{} is in the same range ($45.5\%$ wrong zero-shot, $34.6\%$ five-shot), at lower coverage ($12.4\%$ and $14.6\%$).

\begin{table}[H]
\centering\small
\caption{Confidence failures at confidence $\geq 0.9$. Confident error rate is the fraction of accepted predictions (confidence $\geq 0.9$) that are wrong, with errors / accepted in parentheses; coverage is the fraction of all examples that are accepted. For anomaly rows, FA / miss are confident false alarms and confident misses (counts, not rates). n/a: ATT\&CK is not binary. \colorbox{ours}{Shaded rows}: our model.$^*$ marks metrics computed on a 10\% sample of the corresponding test set and approximated to the full population. }
\label{tab:errors}
\begin{tabular}{@{}l l r r l@{}}
\toprule
Source & Version & Conf.\ error rate (errors / accepted) & Coverage & FA / miss \\
\midrule
\rowcolor{ours} HDFS & \method{} raw & 1.94\% (2189 / 113021) & 0.9827 & 84 / 2105 \\
\rowcolor{ours} HDFS & \method{} fitted & 1.73\% (1913 / 110757) & 0.9630 & 1 / 1912 \\
HDFS & \jev{} (0-shot) & 86.1\% (6593 / 7657) & 0.0666 & 6509 / 84 \\
HDFS & \jev{} (5-shot) & 72.4\% (5819 / 8036) & 0.0699 & 5789 / 30 \\
HDFS & \msdecision{} (0-shot) & 8.18\% (3672 / 44905) & 0.3904 & 3068 / 604 \\
HDFS & \msdecision{} (5-shot) & 5.59\% (4157 / 74330) & 0.6463 & 3337 / 820 \\
HDFS & \oaidecision{} (0-shot)$^*$ & 3.72\% (3265$^*$ / 87875) & 0.7640$^*$ & 2330$^*$ / 935 \\
HDFS & \oaidecision{} (5-shot)$^*$ & 2.97\% (2629$^*$ / 88542) & 0.7698$^*$ & 1590$^*$ / 1039 \\
\midrule
\rowcolor{ours} ATT\&CK (withheld) & \method{} raw & 37.1\% (26 / 70) & 0.3933 & n/a \\
\rowcolor{ours} ATT\&CK (withheld) & \method{} fitted & 33.3\% (6 / 18) & 0.1011 & n/a \\
ATT\&CK (withheld) &\jev{} (0-shot) & 45.5\% (10 / 22) & 0.1236 & n/a \\
ATT\&CK (withheld) & \jev{} (5-shot) & 34.6\% (9 / 26) & 0.1461 & n/a \\
ATT\&CK (withheld) & \msdecision{} (0-shot) & 50.0\% (18 / 36) & 0.2022 & n/a \\
ATT\&CK (withheld) & \msdecision{} (5-shot) & 55.3\% (21 / 38) & 0.2135 & n/a \\
ATT\&CK (withheld) & \oaidecision{} (0-shot) & 58.3\% (7 / 12) & 0.0674 & n/a \\
ATT\&CK (withheld) & \oaidecision{} (5-shot) & 66.7\% (2 / 3) & 0.0169 & n/a \\
\bottomrule
\end{tabular}
\end{table}

\takeaway{Confident errors persist on unseen sources, but in a different form from those of the three hosted decision models. On HDFS, \method{} is confident on almost every session and its confident errors are mostly missed anomalies, whereas those of the three hosted decision models are mostly false alarms. A single fitted temperature removes almost all confident false alarms but leaves most confident misses, and about a third of confident predictions on unseen security classes are still wrong.}

\subsection{Inference latency and cost}
\label{sec:latency-cost}

In this experiment, we compare the latency and per-request cost of \method{} with \jev{}, \msdecision{}, \oaidecision{}, and prompted GPT-5.6 Luna and Sol (Table~\ref{tab:latency-cost}). Latency for \method{} is measured on one NVIDIA A10G GPU at batch size 1. Its cost converts measured latency to AWS \texttt{g5.xlarge} on-demand (\$1.006/hour) and three-year-reserved (\$0.435/hour) pricing. \jev{}'s cost is its metered per-call API cost, and the frontier models' cost applies published per-token rates to measured token counts. All hosted-system latency and cost figures use zero-shot prompts. Hosted latencies include network round-trip time, which local inference avoids.

\textbf{The comparison is tilted toward the hosted systems.} \jev{}, \msdecision{}, \oaidecision{} and GPT-5.6 are served from production inference stacks. As is standard for hosted endpoints, these very likely use continuous batching across many concurrent users, optimized attention kernels, quantization, speculative decoding and hardware tuned for throughput. Their per-call price and latency reflect these optimizations, amortized over the provider's whole traffic. \method{}, by contrast, is served unoptimized: one request at a time, at batch size 1, in standard precision, with no serving framework. Our latency and cost figures are therefore an upper bound for \method{} and a best case for the hosted systems.

\textbf{Even under these conditions, self-hosting is cost-competitive at sustained load.} On the A10G, deciscion APIs (\jev{}, \msdecision{}, \oaidecision{}) faster than \method{} on the A10G. However, only \jev{} and \msdecision{} have lower per-request costs than on-demand \method{} on both tasks; \oaidecision{} is more expensive on both. \method{} becomes the cheapest option with reserved instances, which assume sustained load. GPT-5.6 Luna costs roughly $3$--$8\times$ as much as \method{}, depending on pricing and task, and Sol roughly $60$--$180\times$ as much, which is about two orders of magnitude.

\begin{table}[H]
\centering\small
\caption{Inference latency and cost per 1{,}000 requests. \colorbox{ours}{Shaded rows}: our model. \textbf{Bold}: lowest latency and lowest cost per column.}
\label{tab:latency-cost}
\begin{tabularx}{\textwidth}{@{}Y r r r r@{}}
\toprule
 & \multicolumn{2}{c}{Attack classification} & \multicolumn{2}{c}{Anomaly detection} \\
\cmidrule(lr){2-3}\cmidrule(lr){4-5}
System & Latency & Cost/1,000 & Latency & Cost/1,000 \\
\midrule
\rowcolor{ours} \method{}, A10G on-demand & 1{,}429~ms & \$0.5147 & 924~ms & \$0.2583 \\
\rowcolor{ours} \method{}, A10G 3yr-reserved & 1{,}429~ms & \textbf{\$0.2226} & 924~ms & \textbf{\$0.1117} \\
\jev{} (0-shot) & \textbf{167~ms} & \$0.2527 & \textbf{143~ms} & \$0.2497 \\
\msdecision{} (0-shot) & 675~ms & \$0.3596 & 497~ms & \$0.2279 \\
\oaidecision{} (0-shot) & 453~ms & \$0.8055 & 434~ms & \$0.3849 \\
GPT-5.6 Luna (0-shot) & 1{,}834~ms & \$1.6713 & 1{,}593~ms & \$0.8536 \\
GPT-5.6 Sol (0-shot)& 1{,}836~ms & \$31.6686 & 1{,}814~ms & \$20.0304 \\
\bottomrule
\end{tabularx}
\end{table}
\textbf{A hardware change alone closes most of the latency gap.} We also measured \method{} on one NVIDIA H100 at batch size 1, for anomaly detection only. Latency falls from $924$~ms to about $172$~ms. This is roughly $5\times$ faster and within about $30$~ms of \jev{} ($143$~ms), with no change to the model or serving code. At batch size 1 the H100 costs about \$0.33 per 1{,}000 requests on demand and \$0.14 with three-year reserved pricing. That is more than the A10G, because a single request leaves most of the H100's capacity idle. A standard serving framework with batching would raise utilization on either GPU and lower the cost per request further. We did not run that configuration. However, the hosted systems' advantage comes from serving infrastructure, not from the models themselves, and that infrastructure is equally available to a self-hosted model.

\takeaway{Served without any inference optimization, \method{} is already the cheapest system on reserved instances and is costlier than \jev{} and  \msdecision{} only on demand. Decision APIs are faster on an A10G, but moving \method{} to an H100 brings it within about $30$~ms of \jev{} on anomaly detection and makes it faster than \msdecision{} and \oaidecision{}, with no change to the model or serving code. Luna is roughly $3$--$8\times$ the cost of \method{}, and Sol is about two orders of magnitude more expensive.}

\section{Limitations and Release Scope}
\label{sec:limitations}

\paragraph{Anomaly sources.}
Public anomaly corpora represent specific systems and historical fault distributions.
Familiar templates and correlated incidents can support strong represented-source performance without establishing generalization to unseen failures.
HDFS is a useful source-transfer test, but one excluded source does not characterize all out-of-domain logs.
Its exclusion from our training pipeline also cannot establish its exclusion from the backbone's pretraining data.

\paragraph{Synthetic data.}
The general log data combines public benchmark records with generated questions, generated examples; the ATT\&CK augmentation is generated by GPT-5.5.
Although no HDFS or withheld-sourcetype record is used as a seed or template, the generator may have seen public log corpora, including HDFS, during its own pretraining, and could reproduce similar templates or field conventions.
The generators include models from the same family as the GPT-5.6 baselines, so synthetic data may transfer some of that family's prior knowledge of log formats and ATT\&CK techniques to Lev-3B.

\paragraph{Security classification.}
Technique labels derived from an attack-data repository may be incomplete or ambiguous at the selected observation window.
The 130-way task is conditional on the offered catalog, and a single-label score does not establish complete attack understanding.
Sparse and singleton classes limit the precision of macro-F1 and of per-class probability analysis.
The withheld split is easier than the represented split for most systems evaluated, which likely reflects its technique mix, so withheld-split scores measure transfer to new sourcetypes rather than to harder examples.

\paragraph{Experimental scope.}
Every variant is a single training run, so differences between variants carry no variance estimate, and small differences on the 178-record withheld split correspond to a few records.
We study one 3B backbone; whether a larger or smaller backbone changes transfer or run-to-run variation is not tested.

\paragraph{Release.}
The release includes the base-model revision, adapter and pointer weights, tokenizer and rendering rules, source manifests, split and group identifiers, calibration parameters, and prediction-level outputs.
Data redistribution follows the terms of each source, separately from model and code licensing; where raw log text cannot be redistributed, reproducible builders and hashes identify the inputs.

\section{Conclusion}
\label{sec:conclusion}
We presented Lev-3B, a compact open decision model for log analysis, and the data recipe that produces it.
The main finding is that general log data spanning additional tasks, tied to neither evaluation task, is what makes the model transfer: it offsets most of the interference from adding security supervision and improves results on both excluded sources.
With it, a 3B model outperforms prompted frontier models and all three hosted decision models on anomaly detection, including on an unseen logging system, at a fraction of the serving cost.
Two problems remain.
On HDFS, the model is confident on almost every session, and its confident errors are missed anomalies that a fitted temperature does not fix.
On security classification, it trails prompted frontier models, and its accuracy depends on how many real examples each technique has.
Addressing these likely requires more diverse log sources, more real security data, and calibration that adapts to the target source.
The released checkpoint, manifests, and prediction files are intended to make these comparisons reproducible.

\clearpage
\bibliographystyle{plain}
\bibliography{references}

\clearpage
\appendix

\section{Prediction Artifacts and Access Ledger}
\label{app:artifacts}
\label{app:ledger}

Each saved prediction contains its evaluation ID, parent or incident group, source, gold label, option set and order, full probability vector, predicted option, checkpoint revision, rendering hash, and coverage status, together with raw logits and the applied temperature.
All tables in this paper are generated from these prediction files and the accompanying manifests, not assembled by hand from separate runs.

The release ledger records the raw-data release and checksum for each source, the grouping and split-builder revision, calibration-pool IDs, and the source of every training, development, and calibration record.

\section{Threshold Transfer under Domain Shift}
\label{app:threshold-transfer}

We tested whether a decision threshold selected on the represented sources transfers to HDFS, either unchanged or after aligning it to the unlabeled HDFS score distribution.
The source threshold maximizes the equal-source mean anomaly F1 across the BGL, Spirit, and Thunderbird development sets.
Target-adaptive rules use only the unlabeled HDFS score distribution, and every threshold was frozen before HDFS labels were accessed.
We compare the fixed threshold $0.5$; the unchanged source threshold; mean- and median-percentile transport; three logit-space interpolations between the source and mean-percentile thresholds; and robust median/IQR transport.

\begin{table}[H]
\centering\small
\caption{Anomaly-class F1 (percent) on the 115{,}013-session HDFS population under source-derived and unlabeled-target-adaptive thresholds. Mean OT and median OT transport the mean and median of the three source fractions above the source threshold. Shrink $s$ interpolates in logit space between the source threshold ($s=0$) and mean OT ($s=1$). Med./IQR applies robust location-scale transport in logit space. Bold: best per row.}
\label{tab:threshold-transfer}
{\scriptsize
\setlength{\tabcolsep}{3.4pt}
\begin{tabular}{@{}lrrrrrrrr@{}}
\toprule
Model & Fixed $0.5$ & Source & Mean OT & Median OT & Shrink $.25$ & Shrink $.50$ & Shrink $.75$ & Med./IQR \\
\midrule
\method{} & \textbf{43.32} & 28.37 & 18.20 & 34.58 & 43.08 & 41.02 & 34.42 & 43.02 \\
\jev{}       & 5.65 & 11.85 & 8.97 & \textbf{11.87} & 11.24 & 11.77 & \textbf{11.87} & 8.63 \\
\bottomrule
\end{tabular}
}
\end{table}

For \method{}, no rule improves on the fixed $0.5$ threshold.
The source threshold transferred unchanged lowers F1 from 43.32 to 28.37, and full mean-percentile transport lowers it to 18.20; the mildest interpolation and the robust rule come within 0.3 points of the fixed threshold.
For \jev{}, threshold adaptation roughly doubles F1, but to at most 11.87.
Aligning score distributions moves the operating point but does not fix the mismatch in decision boundaries across systems.
We therefore report HDFS F1 at the fixed $0.5$ threshold in the main text.
\section{ATT\&CK: Synthetic Augmentation and Training Support}
\label{app:provisional}
\label{app:training-support}

\subsection{Split construction}
ATT\&CK records from Splunk's attack-data repository~\citep{splunk2026attackdata} are organized by technique.
The withheld split contains all records from 18 sourcetypes that contribute no training data.
In addition, 30\% of techniques, chosen at random, have no real training records, so some withheld-split techniques are supported in training only by synthetic examples or not at all.
For the remaining techniques, real records are split per technique into train, development, calibration, and test sets, with at least one training record guaranteed before any other split is filled.
Synthetic examples are generated as described in Section~\ref{sec:stage2}.

\subsection{Effect of synthetic augmentation}
Table~\ref{tab:provisional-attack} compares checkpoints trained with real ATT\&CK records only and with synthetic augmentation.
Both are D$\rightarrow$A+T checkpoints without general log data; the synthetic-augmented checkpoint is the D$\rightarrow$A+T row of Table~\ref{tab:attack-ablation}.

\begin{table}[H]
\centering\small
\caption{Effect of synthetic ATT\&CK augmentation (percent). Both Lev rows are D$\rightarrow$A+T, without general log data.}
\label{tab:provisional-attack}
\begin{tabularx}{\textwidth}{@{}Y r r r r@{}}
\toprule
 & \multicolumn{2}{c}{Represented sourcetypes} & \multicolumn{2}{c}{Withheld sourcetypes} \\
\cmidrule(lr){2-3}\cmidrule(lr){4-5}
Checkpoint & Accuracy & Macro-F1 & Accuracy & Macro-F1 \\
\midrule
D$\rightarrow$A+T, real only & 31.3 & 24.7 & 34.8 & 19.7 \\
D$\rightarrow$A+T, real + synthetic & \textbf{35.8} & \textbf{31.7} & 34.3 & 19.8 \\
\bottomrule
\end{tabularx}
\end{table}

Synthetic augmentation improves represented-sourcetype accuracy by 4.5 points and macro-F1 by 7.0 points, but leaves the withheld split essentially unchanged (accuracy $-0.5$, macro-F1 $+0.1$).
Synthetic examples therefore help on familiar sourcetypes but do not by themselves improve transfer; the withheld-split gain of the full recipe comes from general log data (Section~\ref{sec:exp-recipe}).
Both rows are single runs and differ in total training exposure as well as in data content, so the in-domain gain is not attributable to synthetic content alone.

\subsection{Performance by real training support}
Table~\ref{tab:provisional-support} breaks results down by the number of real training records per technique: high ($\geq 10$), medium (2--9), low (1), and zero.
A technique with zero real support may still have synthetic training examples.

\begin{table}[H]
\centering\small
\caption{Accuracy and macro-F1 (percent) by real training support per technique, for \jev{} and the synthetic-augmented D$\rightarrow$A+T checkpoint. Buckets count real training records only.}
\label{tab:provisional-support}
\begin{tabularx}{\textwidth}{@{}l Y r r r r r@{}}
\toprule
 & & & \multicolumn{2}{c}{Accuracy} & \multicolumn{2}{c}{Macro-F1} \\
\cmidrule(lr){4-5}\cmidrule(lr){6-7}
Split & Real support & $n$ & \jev{} & \method{} & \jev{} & \method{} \\
\midrule
Test & High ($\geq10$) & \blank & 28.1 & 40.4 & 36.6 & 43.6 \\
Test & Medium (2--9) & \blank & 19.7 & 31.1 & 18.2 & 31.8 \\
Test & Low (1) & \blank & 37.5 & 37.5 & 33.3 & 35.4 \\
\midrule
Withheld & High ($\geq10$) & \blank & 24.4 & 43.9 & 33.2 & 35.1 \\
Withheld & Medium (2--9) & \blank & 35.3 & 27.9 & 33.8 & 23.5 \\
Withheld & Low (1) & \blank & 47.4 & 26.3 & 28.2 & 24.1 \\
Withheld & Zero (0) & \blank & 22.2 & 11.1 & 20.8 & 8.3 \\
\bottomrule
\end{tabularx}
\end{table}

On the withheld split, Lev's advantage over \jev{} comes entirely from high-support techniques; \jev{} is more accurate in every lower bucket, and much more so on techniques with zero or one real training record.
Lev's performance on a technique therefore depends on how many real examples it saw, and synthetic examples do not substitute for them.

Table~\ref{tab:training-support-real} applies the same idea to the release checkpoint and the frontier models, restricting the withheld split to techniques above a real-support floor.

\begin{table}[H]
\centering\small
\caption{Withheld-sourcetype accuracy / macro-F1 (percent) for the release checkpoint, restricted to techniques with at least the given number of real (non-synthetic) training records. \jev{} uses the same zero-shot predictions as Table~\ref{tab:attack-main}.}
\label{tab:training-support-real}
\begin{tabularx}{\textwidth}{@{}Y r r r r r@{}}
\toprule
Real-support floor & $n$ (of 178) & \method{} & \jev{} (0-shot) & GPT-5.6 Luna & GPT-5.6 Sol \\
\midrule
None & 178 & 42.1 / 29.5 & 32.6 / 28.1 & 51.7 / 42.7 & 51.7 / 45.4 \\
$\geq 5$ (19 techniques) & 117 & 41.9 / 25.3 & 22.2 / 22.6 & 45.3 / 31.9 & 44.4 / 37.0 \\
$\geq 10$ (11 techniques) & 82 & 48.8 / 33.3 & 26.8 / 34.0 & 56.1 / 37.9 & 53.7 / 45.9 \\
$\geq 20$ (4 techniques) & 25 & 52.0 / 50.0 & 28.0 / 38.6 & 60.0 / 54.9 & 64.0 / 56.4 \\
\bottomrule
\end{tabularx}
\end{table}

\method{}'s accuracy rises with the support floor, from 41.9\% at $\geq 5$ to 52.0\% at $\geq 20$, and it is 19.7--24.0 percentage points more accurate than \jev{} at the three nonzero support floors; on macro-F1 it leads \jev{} at $\geq 5$ and $\geq 20$ but trails at $\geq 10$.
The frontier models, which have no training exposure to these sources, also improve as the floor rises (Luna: 45.3\% to 60.0\%), and \method{}'s accuracy gap to them does not shrink (3.4, 7.3, and 8.0 points to Luna).
Techniques with more real training records therefore appear to be easier in general, so this analysis does not show that the frontier gap is caused by limited training exposure.
Only the macro-F1 gap to Luna narrows, from 13.2 points without a floor to 4.9 at $\geq 20$.
The $\geq 20$ row covers four techniques and 25 records and should be read as a directional signal.

\section{Decision and General Log Data Details}
\label{app:data-details}

This section expands the source-access summary in Table~\ref{tab:data} with the provenance and composition of the natural-language decision suite and general log data.
We distinguish a \emph{request}, which contains a shared input state and one or more questions, from a \emph{labeled question}.
Counts describe serialized data before online option augmentation, epoch repetition, or replay; they do not imply that every request has a distinct underlying text or log state.

\subsection{Decision data (D)}
\label{app:decision-v7-details}

The decision data D has two parts: the training partition of Decision-v7, a frozen natural-language decision suite distributed with \kev{}~\citep{palmer2026kev}, and a supplementary set of 1,425 examples that \kev{} generates programmatically.
We describe Decision-v7 first and the supplement at the end of this subsection.

\paragraph{Origin and tasks.}
The public portion of Decision-v7 draws on AG News, Amazon reviews, Banking77, BoolQ, DBpedia14, IMDB, MNLI, SST5, TREC, and Yelp.
\kev{}'s shared decision interface converts each dataset's original annotations into binary propositions, categorical alternatives, or ordinal levels.
The suite also includes programmatically generated policy decisions and compositional rule-following examples.
All of these are natural-language decisions; none contains system logs.

\paragraph{Source distribution.}
The training partition contains 12,576 requests and 15,576 labeled questions (Table~\ref{tab:decision-v7-sources}).
The ten public datasets contribute 1,000 requests each, which together make up 79.5\% of training requests.
Legacy-policy examples contribute 896 requests (7.1\%), and compositional examples contribute 1,680 (13.4\%).
Most requests carry a single labeled question, with two exceptions.
Each AG News request pairs a topic question with two binary topic propositions, and each Yelp request pairs a rating question with a recommendation question.
Equal request counts across public sources therefore do not imply equal question counts.
The request-averaged loss in Eq.~\eqref{eq:loss} accounts for this difference.

\begin{table}[htbp]
\centering\small
\caption{Decision-v7 training composition. Share is the percentage of the 12,576 training requests, rounded to one decimal. Policy and compositional data are generated by \kev{}; the other rows are converted from public NLP datasets.}
\label{tab:decision-v7-sources}
\begin{tabularx}{\textwidth}{@{}l X r r r@{}}
\toprule
Source & Decision content & Requests & Questions & Share (\%) \\
\midrule
AG News & News topic and binary topic propositions & 1,000 & 3,000 & 8.0 \\
Amazon reviews & Five-level product-review rating & 1,000 & 1,000 & 8.0 \\
Banking77 & Banking intent, 77 alternatives & 1,000 & 1,000 & 8.0 \\
BoolQ & Yes/no question answering & 1,000 & 1,000 & 8.0 \\
DBpedia14 & Encyclopedia category, 14 alternatives & 1,000 & 1,000 & 8.0 \\
IMDB & Binary movie-review sentiment & 1,000 & 1,000 & 8.0 \\
MNLI & Entailment, contradiction, or neutral & 1,000 & 1,000 & 8.0 \\
SST5 & Five-level sentence sentiment & 1,000 & 1,000 & 8.0 \\
TREC & Question answer type, six alternatives & 1,000 & 1,000 & 8.0 \\
Yelp & Five-level rating and binary recommendation & 1,000 & 2,000 & 8.0 \\
Legacy policy & Eligibility, thresholds, and policy outcomes & 896 & 896 & 7.1 \\
Compositional & Decisions under combined logical rules & 1,680 & 1,680 & 13.4 \\
\midrule
Total & & 12,576 & 15,576 & 100.0 \\
\bottomrule
\end{tabularx}
\end{table}
\paragraph{Question types and label balance.}
The training questions comprise 5,224 binary (33.5\%), 6,904 categorical (44.3\%), and 3,448 ordinal (22.1\%) questions.
These correspond to \kev{}'s \texttt{noul}, \texttt{choice}, and \texttt{score} interfaces, respectively.
Balancing by source does not balance classes.
BoolQ has 639 positive and 361 negative answers, and IMDB has 492 positive and 508 negative reviews.
Each of the 77 Banking77 intents has between 2 and 26 training examples, and each of the six TREC answer types has between 13 and 240.
These figures are label frequencies in the frozen training sample, not class frequencies in the full upstream datasets.

\paragraph{Supplementary examples and follow-up stage.}
The 1,425 supplementary examples fall into two groups.
The first group contains 900 date-based policy decisions involving return windows, warranty claims, and shipping delays.
The second group contains 525 examples that test whether the model recognizes missing evidence.
In 256 of these, the deciding evidence has been removed, and the target is a uniform distribution over the offered alternatives.
The remaining 269 are intact controls that keep their original labels.
A decision-only follow-up stage trains on these 1,425 examples mixed with 2,000 examples replayed from the Decision-v7 training partition.
The replayed examples add exposure to existing training data, not new source records.



\subsection{General log data: provenance and distribution}
\label{app:general-log-details}

\paragraph{Content and provenance.}
The general log slice L contains 28,740 requests across 35 task definitions, grouped into 14 task families in Table~\ref{tab:general-log-composition}.
Each request contains one labeled question and a task-dependent state: a log line, session or window, build log, incident, or trace-derived context.
The slice combines converted public benchmark records, generated questions or examples, and mechanically derived labels or contrasts.
The task families reuse some underlying logging systems and are therefore not 14 disjoint original log corpora.

LogGLUE-derived tasks cover summary selection, semantic relationships and token categories~\citep{logglueDataset}.
Template classification and Hadoop session tasks use Loghub-derived material~\citep{zhu2023loghub}, while LogQA-style questions use BGL, Spirit, Thunderbird, and OpenSSH contexts.
Other families cover software-build outcomes and failure evidence, microservice faults, root-cause localization, insider-threat campaigns, and questions derived from traces or incident records.
Their sources include LO2v2~\citep{lo2v2Dataset}, Fedora Log Detective~\citep{logdetectiveDataset}, Nezha~\citep{nezhaDataset}, LogChunks~\citep{logchunksDataset}, CoreOps~\citep{coreopsDataset}, CERT~\citep{certDataset}, and RCAEval~\citep{rcaevalDataset}, together with Java build results and AnoMod-derived QA.
Source links and per-record provenance are retained in the accompanying data export.
{
\begin{table}[htbp]
\centering\footnotesize
\caption{General log training data distribution across different task families. }
\label{tab:general-log-composition}
\begin{tabularx}{\textwidth}{@{}p{4cm} X r r r@{}}
\toprule
Task family & Question content & Tasks & Requests & Share (\%) \\
\midrule
LogGLUE-derived & Summaries, semantic relations, tokens, and contrasts & 6 & 7,568 & 26.3 \\
LogQA-style & Answer selection and log-field or template aspects & 5 & 4,769 & 16.6 \\
Trace LogQA & Trace, event, timing, and evidence questions & 1 & 4,000 & 13.9 \\
Template classification & Original and derived log-template choices & 2 & 2,000 & 7.0 \\
Java build results & Build success and failure category & 2 & 2,000 & 7.0 \\
LO2v2 & Fault-injection status and affected service & 2 & 2,000 & 7.0 \\
Fedora Log Detective & Explanatory spans and failure evidence & 2 & 1,652 & 5.7 \\
AnoMod-derived QA & Answers about service scenarios & 1 & 1,000 & 3.5 \\
Nezha & Responsible service, log details, and metric extrema & 3 & 935 & 3.3 \\
LogChunks & Failure-chunk selection and span interpretation & 2 & 913 & 3.2 \\
Hadoop session tasks & Anomaly, fault type, and session details & 4 & 755 & 2.6 \\
CoreOps & Fault classification and remediation & 2 & 586 & 2.0 \\
CERT insider threat & Malicious-campaign identification & 1 & 412 & 1.4 \\
RCAEval-derived QA & Incident details and root-cause service & 2 & 150 & 0.5 \\
\midrule
\multicolumn{2}{@{}l}{Total} & 35 & 28,740 & 100.0 \\
\bottomrule
\end{tabularx}
\end{table}
}

\paragraph{Task and label distribution.}
L contains 24,760 categorical questions (86.2\%) and 3,980 binary questions (13.8\%), with no ordinal questions in this slice.
LogGLUE-derived, LogQA-style, and Trace LogQA requests together account for 56.8\% of L.
The mixture is thus heterogeneous and unequal in size across tasks; its request shares should not be interpreted as the natural prevalence of each kind of operational event.
Class balance is task-specific: the CERT slice has 38 malicious and 374 non-malicious requests, LO2v2 has 983 fault-injected and 17 non-injected requests, and the Java build-success task has 519 successful and 481 failed builds.

\paragraph{Relation to the training mixtures.}
The 28,740-request general log data excludes HDFS logs or Hadoop logs in general.


\end{document}